\documentclass[11pt,a4paper]{article}
\usepackage{acl2020}
\usepackage{latexsym}
\usepackage{url}
\usepackage{textcomp}
\usepackage{graphicx}
\usepackage{array}
\usepackage{dialogue}
\usepackage{xcolor}
\usepackage{booktabs}
\usepackage{subcaption}

\usepackage{times}

\usepackage{adjustbox}
\usepackage[utf8]{inputenc}

\DeclareFixedFont{\ttb}{T1}{txtt}{bx}{n}{12} 
\DeclareFixedFont{\ttm}{T1}{txtt}{m}{n}{12}  

\usepackage{color}
\definecolor{deepblue}{rgb}{0,0,0.5}
\definecolor{deepred}{rgb}{0.6,0,0}
\definecolor{deepgreen}{rgb}{0,0.5,0}
\definecolor{codegreen}{rgb}{0,0.6,0}
\definecolor{codegray}{rgb}{0.5,0.5,0.5}
\definecolor{codepurple}{rgb}{0.58,0,0.82}
\usepackage{listings}

\lstdefinestyle{mystyle}{
    commentstyle=\color{codegreen},
    keywordstyle=\color{magenta},
    stringstyle=\color{codepurple},
    basicstyle=\ttfamily\footnotesize,
    breakatwhitespace=false,         
    breaklines=true,                 
    captionpos=b,                    
    keepspaces=true,                 
    numbersep=5pt,                  
    showspaces=false,                
    showstringspaces=false
    showtabs=false,                  
    tabsize=2,
    frame=lines
}

\newcommand\pythonstyle{\lstset{
language=Python,
basicstyle=\ttm,
otherkeywords={self},             
keywordstyle=\ttb\color{deepblue},
emph={MyClass,__init__},          
emphstyle=\ttb\color{deepred},    
stringstyle=\color{deepgreen},
frame=tb,                         
showstringspaces=false,            %
numbers=left,
stepnumber=1,
}}

\newcommand\blfootnote[1]{%
   \begingroup
   \renewcommand\thefootnote{}\footnote{#1}%
   \addtocounter{footnote}{-1}%
   \endgroup
}

\lstnewenvironment{python}[1][]
{
\pythonstyle
\lstset{#1}
}
{}

\newcommand\pythoninline[1]{{\pythonstyle\lstinline!#1!}}

\usepackage{microtype}

\aclfinalcopy 

\title{ADVISER: A Toolkit for Developing \textit{Multi-modal}, \textit{Multi-domain} and \textit{Socially-engaged} Conversational Agents}

\author{Chia-Yu Li, Daniel Ortega, Dirk V\"ath, Florian Lux, \\ \textbf{Lindsey Vanderlyn, Maximilian Schmidt, Michael Neumann, Moritz V\"olkel,} \\ \textbf{Pavel Denisov, Sabrina Jenne, Zorica Kacarevic and Ngoc Thang Vu*} \\ 
Institute for Natural Language Processing (IMS), University of Stuttgart \\
\texttt{thangvu@ims.uni-stuttgart.de}}

\date{}

\begin{document}
\maketitle
\begin{abstract}
We\blfootnote{* All authors contributed equally.} present ADVISER\footnote{Link to open-source code: \url{https://github.com/DigitalPhonetics/adviser}} - an open-source, multi-domain dialog system toolkit that enables the development of multi-modal (incorporating speech, text and vision), socially-engaged (e.g. emotion recognition, engagement level prediction and backchanneling) conversational agents. The final Python-based implementation of our toolkit is flexible, easy to use, and easy to extend not only for technically experienced users, such as machine learning researchers, but also for less technically experienced users, such as linguists or cognitive scientists, thereby providing a flexible platform for collaborative research. 
\end{abstract}

\section{Introduction}
\label{sec:intro}
Dialog systems or chatbots, both text-based and multi-modal, have received much attention in recent years, with an increasing number of dialog systems in both industrial contexts such as Amazon \textit{Alexa}, Apple \textit{Siri}, Microsoft \textit{Cortana}, Google \textit{Duplex}, \textit{XiaoIce} \cite{zhou2018design} and Furhat\footnote{https://docs.furhat.io}, as well as academia such as \textit{MuMMER} \cite{Foster2016} and \textit{Alana} \cite{Curry2018}. However, open-source toolkits and frameworks for developing such systems are rare, especially for developing multi-modal systems comprised of speech, text, and vision.
Most of the existing toolkits are designed for developing dialog systems focused only on core dialog components, with or without the option to access external speech processing services (\citealp{Bohus2009}; \citealp{Baumann2012}; \citealp{Lison2016}; \citealp{Ultes2017}; \citealp{ortega2019adviser}; \citealp{Lee2019}). 
\\ \indent To the best of our knowledge, there are only two toolkits, proposed in \cite{Foster2016} and \cite{Bohus2017}, that support developing dialog agents using multi-modal processing and social signals \cite{SSI}. Both provide a decent platform for building systems, however, to the best of our knowledge, the former is not open-source, and the latter is based on the .NET platform, which could be less convenient for non-technical users such as linguists and cognitive scientists, who play an important role in dialog research. 
\\ \indent In this paper, we introduce a new version of ADVISER - previously a text-based, multi-domain dialog system toolkit \cite{ortega2019adviser} - that supports multi-modal dialogs, including speech, text and vision information processing. This provides a new option for building dialog systems that is open-source and Python-based for easy use and fast prototyping. The toolkit is designed in such a way that it is modular, flexible, transparent, and user-friendly for both technically experienced and less technically experienced users.
\\ \indent Furthermore, we add novel features to ADVISER, allowing it to process social signals and to incorporate them into the dialog flow. We believe that these features will be key to developing human-like dialog systems because it is well-known that social signals, such as emotional states and engagement levels, play an important role in human computer interaction \cite{mctear2016conversational}. However in contrast to open-ended dialog systems \cite{ElizaWeizenbaum66}, our toolkit focuses on task-oriented applications \cite{GUS}, such as searching for a lecturer at the university \cite{ortega2019adviser}. 
The purpose we envision for dialog systems developed using our toolkit is not the same as the objective of a social chatbot such as \textit{XiaoIce} \cite{zhou2018design}. Rather than promoting ``an AI companion with an emotional connection to satisfy the human need for communication, affection, and social belonging'' \cite{zhou2018design}, ADVISER helps develop dialog systems that support users in efficiently fulfilling concrete goals, while at the same time considering social signals such as emotional states and engagement levels so as to remain friendly and likeable.

\section{Objectives}
\label{sec:design}
The main objective of this work is to develop a multi-domain dialog system toolkit that allows for multi-modal information processing and that provides different modules for extracting social signals such as emotional states and for integrating them into the decision making process. The toolkit should be easy to use and extend for users of all levels of technical experience, providing a flexible collaborative research platform.

\subsection{Toolkit Design}
We extend and substantially modify our previous, text-based dialog system toolkit \cite{ortega2019adviser} while following the same design choices. This means that our toolkit is meant to optimize the following four criteria: \textit{Modularity}, \textit{Flexibility}, \textit{Transparency} and \textit{User-friendliness at different levels}. This is accomplished by decomposing the dialog system into independent modules (services), which in turn are either rule-based, machine learning-based or both. These services can easily be combined in different orders/architectures, providing users with flexible options to design new dialog architectures. 

\subsection{Challenges \& Proposed Solutions}
\paragraph{Multi-modality} The main challenges in handling multi-modality are a) the design of a synchronization infrastructure and b) the large range of different latencies from different modalities. 
To alleviate the former, we use the publisher/subscriber software pattern presented in  section \ref{sec:software} to synchronize signals coming from different sources.
This software pattern also allows for services to run in a distributed manner. By assigning computationally heavy tasks such as speech recognition and speech synthesis to a more powerful computing node, it is possible to reduce differences in latency when processing different modalities, therefore achieving more natural interactions.

\paragraph{Socially-Engaged Systems}
Determining the ideal scope of a socially-engaged dialog system is a complex issue, that is which information should be extracted from users and how the system can best react to these signals.
Here we focus on two major social signals: emotional states and engagement levels (see section \ref{sec:emotion_engagement}), and maintain an internal user state to track them over the course of a dialog. Note that the toolkit is designed in such a way that any social signal could be extracted and leveraged in the dialog manager. 
In order to react to social signals extracted from the user, we provide an initial affective policy module (see section \ref{sub:policies}) and an initial affective NLG module (see section \ref{sub:NLG}), which could be easily extended to more sophisticated behavior.
Furthermore, we provide a backchanneling module that enables the dialog system to give feedback to users during conversations. Utilizing these features could lead to increased trust and enhance the impression of an empathetic system.

\section{Functionalities}
\label{sec:functionalities}
\subsection{Social Signal Processing}
\label{sec:emotion_engagement}
\begin{figure}
    \centering
    \includegraphics[width=0.5\textwidth]{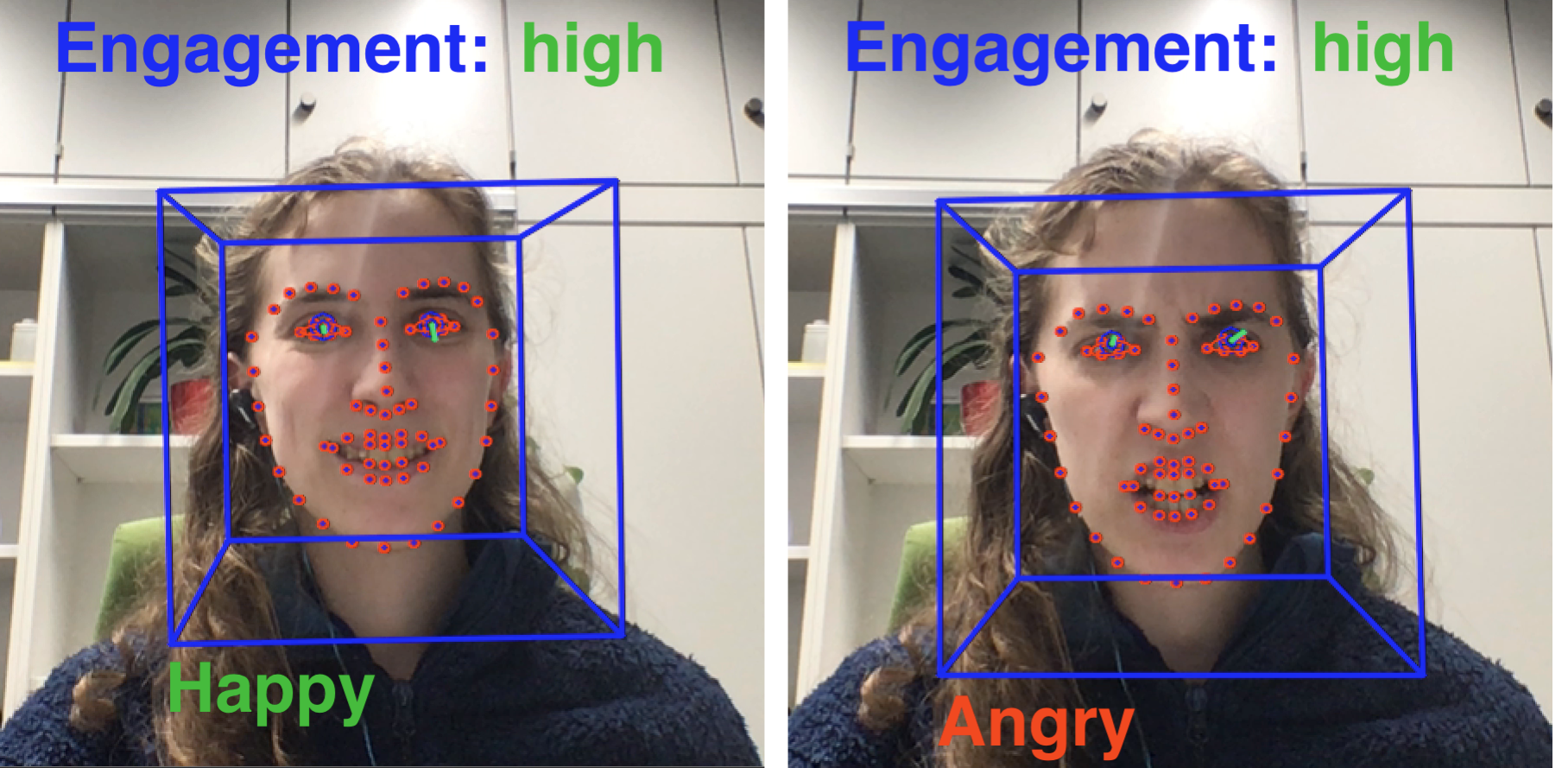}
    \caption{Tracking emotion states and engagement levels using multi-modal information.}
    \label{fig:face_landmarks}
\end{figure}

We present the three modules of ADVISER for processing social signals: (a) emotion recognition, (b) engagement level prediction, and (c) backchanneling. Figure \ref{fig:face_landmarks} illustrates an example of our system tracking emotion states and engagement levels. 
\paragraph{Multi-modal Emotion Recognition} 
For recognizing a user's emotional state, all three available modalities -- text, audio, and vision -- can potentially be exploited, as they can deliver complementary information~\cite{zeng2009survey}. Therefore, the emotion recognition module can subscribe to the particular input streams of interest (see section~\ref{sec:software} for details) and apply emotion prediction either in a time-continuous fashion or discretely per turn.
\\ \indent In our example implementation in the toolkit, we integrate speech emotion recognition, i.e. using the acoustic signal as features.
Based on the work presented in~\cite{neumann2017attentive} we use log Mel filterbank coefficients as input to convolutional neural networks (CNNs). For the sake of modularity, three separate models are employed for predicting different types of labels: (a) basic emotions \{angry, happy, neutral, sad\}, (b) arousal levels \{low, medium, high\}, and (c) valence levels \{negative, neutral, positive\}. The models are trained on the IEMOCAP dataset~\cite{busso2008iemocap}. 
The output of the emotion recognition module consists of three predictions per user turn, which can then be used by the user state tracker (see section~\ref{sec:state_tracking}).
For future releases, we plan to incorporate multiple training datasets as well as visual features.

\paragraph{Engagement Level Prediction} 
User engagement is closely related to states such as boredom and level of interest, with implications for user satisfaction and task success (\citealp{Forbes-Riley2012}; \citealp{Schuller2009}). In ADVISER, we assume that eye activity serves as an indicator of various mental states (\citealp{Schuller2009}; \citealp{Niu2018}) and implement a gaze tracker that monitors the user's direction of focus via webcam. 
\\ \indent Using OpenFace 2.2.0, a toolkit for facial behavior analysis \cite{Baltrusaitis2018}, we extract the features \textit{gaze\_angle\_x} and \textit{gaze\_angle\_y}, which capture left-right and up-down eye movement, for each frame and compute the deviation from the central point of the screen. If the deviation exceeds a certain threshold for a certain number of seconds, the user is assumed to look away from the screen, thereby disengaging. Thus, the output of our engagement level prediction module is the binary decision \{looking, not looking\}. Both the spatial and temporal sensitivity can be adjusted, such that developers have the option to decide \textit{how far} and \textit{how long} the user's gaze can stray from the central point until they are considered to be disengaged. In an adaptive system, this information could be used to select re-engagement strategies, e.g. using an affective template (see section \ref{sub:NLG}).

\paragraph{Backchanneling} 
In a conversation, a backchannel (BC) is a soft interjection from the listener to the speaker, with the purpose of signaling acknowledgment or reacting to what was just uttered. Backchannels contribute to a successful conversation flow \cite{clark2004}. Therefore, we add an acoustic backchannel module to create a more human-like dialog experience.
For backchannel prediction, we extract 13 Mel-frequency-cepstral coefficients from the user's speech signal, which form the input to the convolutional neural network based on \citet{OrtegaBC2020}. The model assigns one of three categories from the proactive backchanneling theory \cite{goodwin1986} to each user utterance \{no-backchannel,  backchannel-continuer and backchannel-assessment\}.
The predicted category is used to add the backchannel realization, such as {\it Right} or {\it Uh-huh}, to the next system response.

\subsection{Speech Processing}
\paragraph{Automatic Speech Recognition (ASR)}
The speech recognition module receives a speech signal as input, which can come from an internal or external microphone, and outputs decoded text. The specific realization of ASR can be interchanged or adapted, for example for new languages or different ASR methods. 
We provide an end-to-end ASR model for English based on the Transformer neural network architecture. We use the end-to-end speech processing toolkit \mbox{ESPnet}~\cite{watanabe2018espnet} and the IMS-speech English multi-dataset recipe \cite{denisov2019ims}, updated to match the LibriSpeech Transformer-based system in ESPnet \cite{karita2019comparative} and to include more training data. Training data comprises the LibriSpeech, Switchboard, TED-LIUM~3, AMI, WSJ, Common~Voice~3, SWC, VoxForge and \mbox{M-AILABS} datasets with a total amount of 3249 hours. As input features, \mbox{80-dimensional} log Mel filterbank coefficients are used. Output of the ASR model is a sequence of subword units, which include single characters as well as combinations of several characters, making the model lexicon independent. 

\paragraph{Speech Synthesis}
For ADVISER's voice output, we use the ESPnet-TTS toolkit~\cite{hayashi2019espnettts}, which is an extension of the ESPnet toolkit mentioned above. We use FastSpeech as the synthesis model speeding up mel-spectrogram generation by a factor of 270 and voice generation by a factor of 38 compared to autoregressive Transformer TTS~\cite{ren2019fastspeech}. We use a Parallel WaveGAN~\cite{yamamoto2020parallel} to generate waveforms that is computationally efficient and achieves a high mean opinion score of 4.16. The FastSpeech and WaveGAN models were trained with 24 hours of the LJSpeech dataset from a single speaker ~\cite{ljspeech} and are capable of generating voice output in real-time when using a GPU. The synthesis can run on any device in a distributed system. Additionally, we optimize the synthesizer for abbreviations, such as {\it Prof., Univ., IMS, NLP, ECTS} and {\it PhD}, as well as for German proper names, such as street names. These optimizations can be easily extended.  

\paragraph{Turn Taking} 
To make interacting with the system more natural, we use a naive end-of-utterance detection. Users indicate the start of their turn by pressing a hotkey, so they can choose to pause the interaction.  The highest absolute peak of each recording chunk is then compared with a predefined threshold. If a certain number of sequential chunks do not peak above the threshold, the recording stops. We are currenlty in the process of planning more sophisticated turn taking models, such as \citet{turn-taking}.

\subsection{Natural Language Understanding}  
The natural language understanding (NLU) unit parses the textual user input \cite{de2008spoken} - or the output from the speech recognition system - and extracts the user action type, generally referred to as intent in goal-oriented dialog systems (e.g. {\it Inform} and {\it Request}), as well as the corresponding slots and values.
The domain-independent, rule-based NLU presented in \citet{ortega2019adviser} is integrated into ADVISER and adapted to the new domains presented in section \ref{sec:usecases}.

\subsection{State Tracking} 
\label{sec:state_tracking}
\paragraph{Belief State Tracking (BST):}  
The BST tracks the history of user informs and the user action types, requests, with one BST entry per turn. This information is stored in a dictionary structure that is built up, as the user provides more details and the system has a better understanding of user intent.

\paragraph{User State Tracking (UST):}  
Similar to the BST, the UST tracks the history of the user's state over the course of a dialog, with one entry per turn. In the current implementation, the user state consists of the user's engagement level, valence, arousal, and emotion category (details in section \ref{sec:emotion_engagement}). 

\subsection{Dialog Policies}
\label{sub:policies}

\paragraph{Policies} 
To determine the correct system action, we provide three types of policy services: a handcrafted and a reinforcement learning policy for finding entities from a database \cite{ortega2019adviser}, as well as a handcrafted policy for looking up information through an API call. Both handcrafted policies use a series of rules to help the user find a single entity or, once an entity has been found (or directly provided by the user), find information about that entity. 
The reinforcement learning (RL) policy's action-value function is approximated by a neural network which outputs a value for each possible system action, given the vectorized representation of a turn's belief state as input.
The neural network is constructed as proposed in \citet{vath2019combine} following a duelling architecture \cite{wang2015dueling}. It consists of two separate calculation streams, each with its own layers, where the final layer yields the action-value function. 
For off-policy batch-training, we make use of prioritized experience replay \cite{schaul2015prioritized}.

\paragraph{Affective Policy} 
In addition, we have also implemented a rule-based affective policy service that can be used to determine the system's emotional response. As this policy is domain-agnostic, predicting the next system \textit{emotion} output rather than the next system \textit{action}, it can be used alongside any of the previously mentioned policies.

\paragraph{User Simulator} 
To support automatic evaluation and to train the RL policy, we provide a user simulator service outputting at the user acts level. As we are concerned with task-oriented dialogs here, the user simulator has an agenda-based \cite{schatzmann2007agenda} architecture and is randomly assigned a goal at the beginning of the dialog. Each turn, it then works to first respond to the system utterance, and then after to fulfill its own goal. When the system utterance also works toward fulfilling the user goal, the RL policy is rewarded by achieving a shorter total dialog turn count \cite{ortega2019adviser}.  

\subsection{External Information Resources}
ADVISER supports three options to access information from external information sources. In addition to being able to query information from SQL-based databases, we add two new options that includes querying information via APIs and from knowledge bases (e.g. Wikidata \citep{wikidata}).
For example, when a user asks a simple question - \textit{Where was Dirk Nowitzki born?}, our pretrained neural network predicts the topic entity - \textit{Dirk Nowitzki} - and the relation - \textit{place of birth}. Then, the answer is automatically looked up using Wikidata's SPARQL endpoint.

\subsection{Natural Language Generation (NLG)}
\label{sub:NLG}

In the NLG service, the semantic representation of the system act is transformed into natural language.
ADVISER currently uses a template-based approach to NLG in which each possible system act is mapped to exactly one utterance.
A special syntax using placeholders reduces the number of templates needed and accounts for correct morphological inflections \cite{ortega2019adviser}.
Additionally, we developed an affective NLG service, which allows for different templates to be used depending on the user's emotional state. This enables a more sensitive/adaptive system. For example, if the user is sad and the system does not understand the user's input, it might try to establish common ground to prevent their mood from getting worse due to the bad news. An example response would be ``As much as I would love to help, I am a bit confused'' rather than the more neutral ``Sorry I am a bit confused''.
One set of NLG templates can be specified for each possible emotional state.
At runtime, the utterance is then generated from the template associated with the current system emotion and system action.

\section{Software Architecture}  
\label{sec:software}

\begin{figure}
    \centering
    \includegraphics[width=0.5\textwidth]{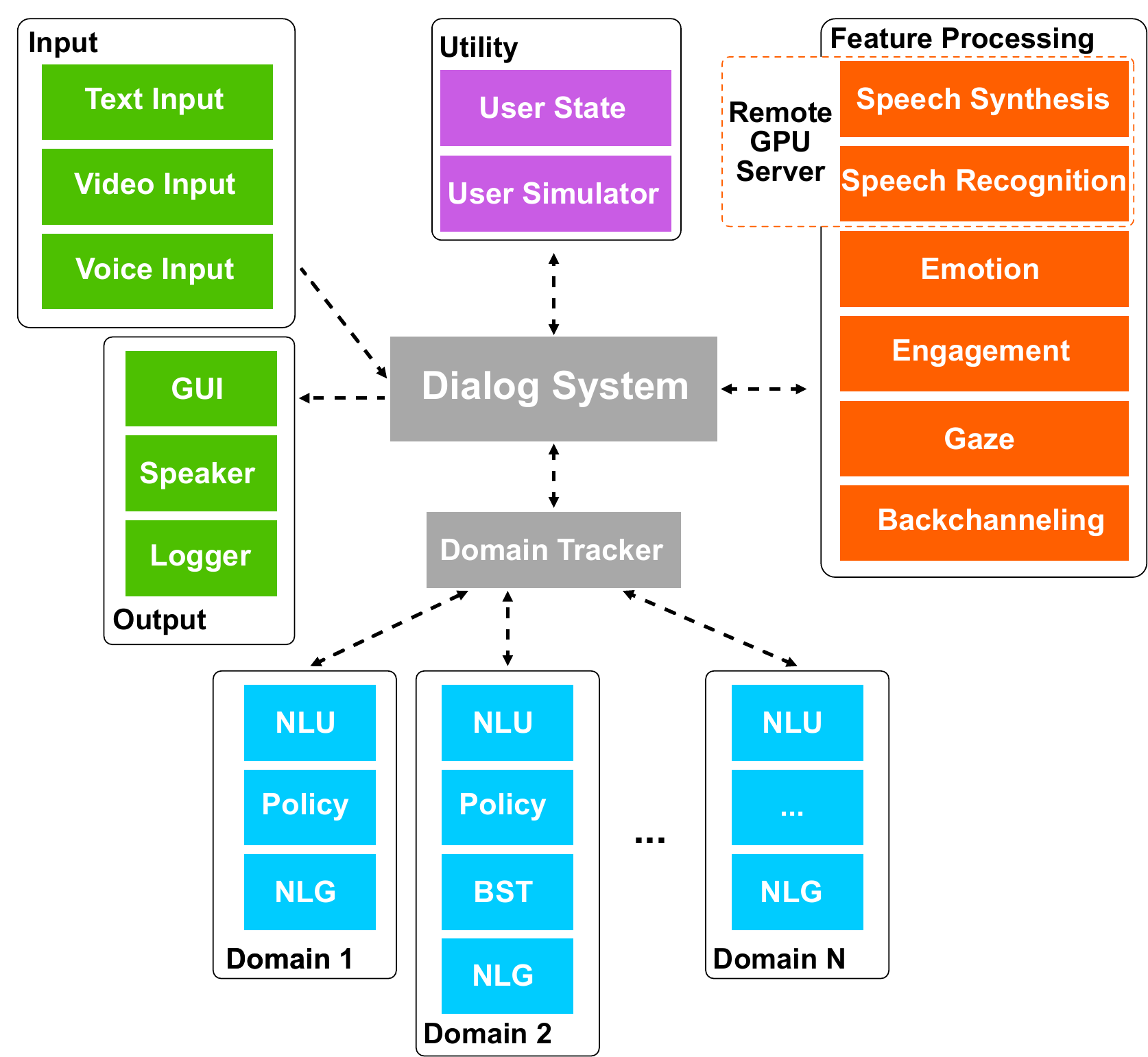}
    \caption{Example ADVISER toolkit configuration: Grey represents backend components, blue represents domain-specific services, and all other colors represent domain-agnostic services. Two components are run remotely.}
    \label{fig:sys_overview}
\end{figure}

\subsection{Dialog as a Collection of Services}
To allow for maximum flexibility in combining and reusing components, we consider a dialog system as a group of services which communicate asynchronously by publishing/subscribing to certain topics. A service is called as soon as at least one message for all its subscribed topics is received and may additionally publish to one or more topics. 
Services can elect to receive the most recent message for a topic (e.g. up-to-date belief state) or a list of all messages for that topic since the last service call (e.g. a list of video frames). Constructing a dialog system in this way allows us to break free from a pipeline architecture. Each step in the dialog process is represented by one or more services which can operate in parallel or sequentially. For example, tasks like video and speech capture may be performed and processed in parallel before being synchronized by a user state tracking module subscribing to input from both sources.
Figure~\ref{fig:sys_overview} illustrates the system architecture.
For debugging purposes, we provide a utility to draw the dialog graph, showing the information flow between services, including remote services, and any inconsistencies in publish/subscribe connections.

\subsection{Support for Distributed Systems}
Services are location-transparent and may thus be distributed across multiple machines. A central dialog system discovers local and remote services and provides synchronization guarantees for dialog initialization and termination.
Distribution of services enables, for instance, a more powerful computer to handle tasks such as real-time text-to-speech generation (see Figure \ref{fig:sys_overview}). This is particularly helpful when multiple resource-heavy tasks are combined into a single dialog system.

\subsection{Support for Multi-Domain Systems}
In addition to providing multi-modal support, the publish/subscribe framework also allows for multi-domain support by providing a structure which enables arbitrary branching and rejoining of graph structures. When a service is created, users simply specify which domain(s) it should publish/subscribe to. This, in combination with a domain tracking service, allows for seamless integration of domain-agnostic services (such as speech input/output) and domain-specific services (such as NLU/NLG for the lecturers domain).

\section{Example Use Cases}
\label{sec:usecases}
\subsection{Example Domains}

We provide several example domains to demonstrate ADVISER's functionalities. 
Databases for lecturers and courses at the Institute for Natural Language Processing (IMS), which we used in the previous version of ADVISER, were adapted to the new system architecture.
As example APIs, we implemented a weather domain that makes calls to the OpenWeatherMap API\footnote{\url{https://openweathermap.org/api}} and a mensa domain for gathering information from the dining hall at the university of Stuttgart.
Note that affective templates were only added to the lecturers and mensa domain.
All domains can be used within the same dialog, simply by switching the topic.

\subsection{An Example Implementation} 

Our toolkit allows for easy creation of a dialog system within a few lines of code as follows.

\begin{lstlisting}[language=python][h!]
# domains
weather = WeatherDomain()
mensa = MensaDomain()
# create system from desired services
ds = DialogSystem(services=[
    DomainTracker(domains=[mensa, weather]),
    SpeechRecorder(),
    VideoInput(),
    EngagementTracker(),
    EmotionTracker(),
    WeatherNLU(domain=weather),
    MensaNLU(domain=mensa),
    ...,
    SpeechOutputPlayer()])

ds.run_dialog(...)
\end{lstlisting}

As a first step, a dialog system object is initialized, which is responsible for coordinating the initialization and graceful termination of all dialog services.
Talking about multiple domains in one dialog is enabled by creating a simple keyword-based domain tracker which is introduced as the first argument to the dialog system.
To make the dialog multi-modal, speech and vision modules are introduced next, along with modules to extract engagement and emotion. So far, all of these modules are domain-agnostic and can be used as shared resources between all domains. 
Next, domain-dependent services such as NLUs, BSTs and NLGs for weather and mensa, are added. 
The following shows an example dialog. 

\vspace{.2cm}
\begin{dialogue}
\footnotesize
\speak{System} Hello, please let me know how I can help you, I can discuss the following domains: Mensa Food and Weather.
\speak{User}I could have something to eat. What does the mensa offer today?
\speak{System} What type of dish are you looking for?
\speak{User}I would like a main dish.
\speak{System} Should the meal be vegan?
\speak{User}Yes.
\speak{System} The meal mediterranean Ebly wheat is served today, is a main dish and is vegan.
\speak{User}Okay, cool, I will go there now! What is the weather like?
\speak{System} The weather in Stuttgart on January 28 at 3 PM is 3 degrees celsius with light snow.
\speak{User}Thank you, ADVISER, good bye!
\speak{System} Thank you, good bye.
\end{dialogue}

\section{Related Work} 
\label{sec:relatedWork}

Other tools for building dialog systems include \textit{ConvLab} \cite{Lee2019}, an open-source, text-based dialog system platform that supports both pipelined architectures and an end-to-end neural architecture. \textit{ConvLab} also provides reusable components and supports multi-domain settings. Other systems are largely text-based, but offer the incorporation of external speech components. \textit{InproTK} \cite{Baumann2012}, for instance, in which modules communicate by networks via configuration files, uses ASR based on Sphinx-4 and synthesis based on MaryTTS. Similarly, \textit{RavenClaw} \cite{Bohus2009} provides a framework for creating dialog managers; ASR and synthesis components can be supplied, for example, by connecting to Sphinx and Kalliope. \textit{OpenDial} \cite{Lison2016} relies on probabilistic rules and provides options to connect to speech components such as Sphinx. Multi-domain dialog toolkit - \textit{PyDial} \cite{Ultes2017} supports connection to DialPort.

As mentioned in the introduction, Microsoft Research's \textit{\textbackslash psi} is an open and extensible platform that supports the development of multi-modal AI systems \cite{Bohus2017}. It further offers audio and visual processing, such as speech recognition and face tracking, as well as output, such as synthesis and avatar rendering. And the \textit{MuMMER} (multi-modal Mall Entertainment Robot) project \cite{Foster2016} is based on the SoftBank Robotics \textit{Pepper} platform, and thereby comprises processing of audio-, visual- and social signals, with the aim to develop a socially engaging robot that can be deployed in public spaces. 

\section{Conclusions}
\label{sec:conclusions}
We introduce ADVISER -- an open-source, multi-domain dialog system toolkit that allows users to easily develop multi-modal and socially-engaged conversational agents. We provide a large variety of functionalities, ranging from speech processing to core dialog system capabilities and social signal processing.
With this toolkit, we hope to provide a flexible platform for collaborative research in multi-domain, multi-modal, socially-engaged conversational agents.

\bibliography{main}
\bibliographystyle{acl_natbib}

\end{document}